*Vilisov V.Ya.*

# ROBOT TRAINING UNDER CONDITIONS OF INCOMPLETE INFORMATION

*Energia IT LLC, Korolyov, vvib@yandex.ru*

The development of the author's works about adaptive algorithms of teaching the robotic systems with the help of operator is described here. An operator is assumed to be an experience decision-maker and sane carrier of a target which the robotic system needs to achieve. The works characteristic is that the robotic system's behavior is not specified a priori (as standard) but is formed adaptively based on the information about the situation and decisions made by a decision-maker. In this scheme the robotic system and the decision-maker can cooperate in the normal operation mode of the robotic system or in the time sharing mode with the possibility to plan actively the experiment on the robotic system. If the adaptive scheme is chosen, there are teaching stages and operating stages of the robotic system. At that the decision-maker can act slowly having the possibility to weigh the decision made. This way allows the robotic system reacting flexibly by switching between preset models and respond to the environment instability.

The data integrity about the environment condition and about target preferences of an operator plays a very important role in robotic system work. The effective work of the robotic system depends on the effective settings of a preference model of the robotic system based on the the decision-maker's decisions and on the effective control. The influence of settings and control factors on the index of effectiveness of the robotic system is subject of this work. The uncertainty may be caused by the data flow limitation received by the operator on the stage of the model setting.

**Introduction**

The conditions of the robotic systems operation during implementation of their functions in different application fields may be different – from manipulation to autonomous problem solving and necessary interaction in a group. At that, the role of a human, operator and/or decision-maker, with a priori or current definition of targets may vary over a wide range.

Hereinafter speaking about robotic systems we will also refer to a person who directly or indirectly participates in the work of the robotic system and is a part of it.

The most part of contexts in which the robotic systems operation is considered to have design character, i.e. there are target and problem, the solving of which the robotic system task providing the quality defines by the metric. At that the access time as a rule is final and is part of a problem limitation. In this variant the robotic system achieves its goal in one operation. Such problems are one-time problems, unique or rare and each solving is different from the others. For example, the robotic system work on another planet, undersea lifting operations etc.

In some cases the robotic system work is a multiple repeat of some sets of problems. In this case the effectiveness is evaluated based on lots of repeated operations, executed by the robotic system. For example, cleaning robots, antimine robots, fire-fighting robots, warrior-robots etc.

The difference between these two variants is that in the first case the decision-maker shall assure the quality of work within the frame of a single operation by interfering from time to time and adjusting the targets, criteria, problem limitation. In the second case the decision-maker may not interfere in a single operation but make adjustments in the following operations based on the observations. The issues described herein refer to the second case.

In the most of areas where the robotic systems are applied the ideal situation is when a single robot or group of robots meets a target in the autonomous mode. But it is not always possible due to the peculiarities of the application field with uncertain environment and limited capacity of the robotic system. The problems of the autonomous work of the robotic systems are described in many works [1, 2] and it comes from not only from the limitation of the power sources but from the environment dynamics, instability of target conditions and other limitations [3].

The author's approach [4, 5] is based on providing the maximum relevancy of the target preferences formalized and used by on-board system of the robotic system for the current control, and provides the high level of the robotic system autonomy which means the effectiveness of the problem solving.

The variants of control strategies [3] (sole, hierarchical, group, gregarious) and control mode (remote, supervisory) are not the subject of this work. The algorithm computational complexity and data exchange inside the robotic system and with an operator (decision-maker) are considered as well. We will study the peculiarities of solution algorithms for allocation problems and their most important characteristics.

The outstanding characteristics of the problem group under consideration and their respective models are the following:

1. The robotic system shall solve the allocation problem with some rate (in random time, as possible).
2. The environment of the robotic system generates situations requiring the decision making (SRDM) which means that the allocation problem between robots shall be solved. The robotic system may act as an initiator of a problem solving when some conditions are met. For example, when the power source level reaches the limit.
3. The robotic system operation effectiveness can not be set a priori with a single scale value for the whole planning period of the robotic system operation [3, 6]. The necessary index of the effectiveness can not be detected and formalized a priori, i.e. on the design stage, adjustment stage and stage of preparation to solve the problem.
4. The control environment may be unstable. It means that not only the limitations but the content of variables and target preferences of a decision-maker may change. At that in spite of the fact that the robotic system works in favour of a decision-maker, the instability may come from it.

**Problem definition**

The allocation problems comprise the essential part of the problems to solve by a separate robot or a group of robots [3, 6].

We will consider the allocation problem or some limited resource in the group of robots as the main context.

The problems of the similar structure arise during control of separate robot, for example, cleaning robots or transport robots which transport cargos along the road network or handle loading operation in a warehouse [3].

This type of problems is structurally represented by the problems of allocation (AP) [3, 6]. Lets examine the most general case – transportation problems (TP) noting that AP is an especial case of transportation problem. The transportation problems can be represented by a wide range of applications, for example, digging-robots which execute the transportation and logistic operations in a network of underground excavations or in space expeditions [3].

As AP is an especial case of TP which is again an especial case of linear programming problems (LPP), then the peculiarities of solving LPP [4] will suit for TP during robotic system control. But the structure of TP determines a number of peculiarities typical only for TP which has to be considered while realizing the control algorithm of the robotic system.

Let's examine the definition and the methods of solving TP in the adaptive form permitting to consider the described peculiarities (1-4) which arise during robotic system control.

To solve TP means to find a set of item values $x_{ij}$ of the source number matrix $X = \|x_{ij}\|_{mn}$ (in the classical TP [7] the resources are considered to be the similar goods transported from points of departure to destination points). The matrix variables $x_{ij}$ is a transportation plan and in the allocation problems – target table in which the variables can posses only Boolean values and in the context of allocation of problems in a group of robots means the $i$-problem assignment to $j$-robot. If we examine the cases when every robot of a group is able to solve several problems (for example, to hit several targets), then the situation will correspond to general TP with discrete variables.

The classical [7] criteria of the plan optimality is the minimum integrated cost of all transportations. The initial data is generally the known vector $\bar{a} = [a_1 \quad a_2 \quad \cdots \quad a_m]^T$ of stock size in a point of departure, where T – is a symbol of transportation; vector $\bar{b} = [b_1 \quad b_2 \quad \ldots \quad b_n]^T$ of a demand rate for each destination point. Matrix $C = \|c_{ij}\|_{mn}$ of the transportation cost of goods from $i$-point of departure to $j$ −destination point is generally known. For the robotic system $c_{ij}$ the efficiency of a problem solving of $i$-type by $j$-robot makes sense. At that, it is consider that the general effect is scaled and has an additive structure, i.e. is formed as the sum of separate effects. Traditionally [7], TP are solved based on the criteria of maximizing some effect or based on the criteria of minimizing some expenses like time consumption or other costs. Bur such a rigid determination of what to be minimized or maximized is important only when TP is solved according to the traditional (standard) scheme. At that, the optimal solving will be optimal accurate to the criterion of optimality. As far as the reality in the vast majority of cases is multicrtiterion for almost all applications, then the real effectiveness of solving (transportation plan, target table) may be far away from the level which satisfies a decision-maker acting as the target definer, or a person who knows the desired level of the robotic system effectiveness in current time. The basic contradiction arises here and this contradiction restrains the effective use of TP model (and other similar) connected to the application multicrtiterion demand and single criterion possibilities of the traditional models. One of the methods to overcome the contradiction is to change the standard scheme for model making to adaptive one [4, 5]. The clue of the adaptive scheme is finding some generalized scaled target function which would approximate the whole vector of target functions of the decision-maker (obvious and unobvious) and send it to the robotic system to execute. Then the decision-maker shall always be satisfied with the robotic system effectiveness at the acceptable level of approximation because the target function (TF) used during the transportation plan construction in any time is a formalized image of its personal criteria preferences.

We go into mathematic definition of TP in a standard form and then we will show the features of its adaptive variant.

TP as one of the variants of the linear programming problem has been historically isolated into an independent group due to its specific structure which allows solving it effectively with the help of specially designed methods based on hand calculations. But modern programs and computer tools allows using the standard solving for the linear programming problems by modifying TP into LPP. Now we will show how to present the initial definition of TP as a standard LPP. This fact allows implementing the adaptive variant of TP based on the similar means of LLP [4].

Let's consider that some initial (a priori) variant of elements $c_{ij}$ of of the game matrix is found in the planning algorithm of the robotic system. These initial evaluations will be specified in the process of adaptation of TP to the target preferences of LPP. The general scheme of the adaptive variant realization of TP (ATP) in the robotic system shall consist on the following stages.

1. Situations represented as set of two vectors $\{\bar{a}, \bar{b}\}$ and requiring the decision making, i.e. planning of transportation and assignments, are recognized by the robotic system which solves TP based on the values of the game matrix $C$. The result of the solving is the matrix $X$. Such a problem we will call the direct TP (DTP).

2. The found solving $X$ is implemented and the decision-maker observes the $L(X)$ effect.

3. The decision-maker evaluates the decision made $q \in \{0; 1\}$, if it's good or bad (i.e. optimal or nonoptimal in his opinion), based on the results of observation of the set $\{\bar{a}, \bar{b}, X, L(X)\}$.

4. Based on the set of data $\{\bar{a}, \bar{b}, X, q\}$ the values of the game matrix $C$ are specified. The values of the game matrix then become current for the next planning step. The elements of the game matrix are specified by solving the inverse TP (ITP), the algorithms of which realize the adaptation mechanism (feedback) allowing to support the actual target function of the robotic system and similar to the current target preferences of the decision-maker.

In such a manner the sequence of described four steps is the iterative procedure in which DTP and ITP are solved by turn. But if there is a reason to consider that the environment and preferences of the decision-maker in some time stays unchanged, then ITP (the main function of which is to provide for the current adequacy of the target function of the robotic system to the target function of the decision-maker) may not be solved. The only thing left is to execute the planning by solving DTP.

Example of the direct and inverse TP.

Target function of TP has the following view:

$$L(X) = \sum_{i=1}^{m} \sum_{j=1}^{n} c_{ij} x_{ij} \qquad (1)$$

SRDM is defined by the set of two vectors $\{\bar{a}, \bar{b}\}$ which shall meet the following limitations for the balanced TP:

$$\sum_{j=1}^{n} x_{ij} = a_i, \quad i = 1, \dots, m; \qquad (2)$$

$$\sum_{i=1}^{m} x_{ij} = b_i, \quad j = 1, \dots, n; \qquad (3)$$

$$x_{ij} \geq 0, \quad i = 1, \dots, m; \quad j = 1, \dots, n. \qquad (4)$$

If the elements of the game matrix have the meaning to win, then the TP criterion has the following view:

$$X_{opt} \to \arg\max_{X} L(X) \qquad (5)$$

In such a manner the relations (1) - (5) are definition of the direct TP realized in point 1 of algorithm. The result of it is the optimal transportation plan (allocation or definition of problems in the robotic system). Here we consider that all elements in different moments of planning are exactly known. The only element of the problem requiring the precision with the help of solving the inverse TP (see p.4 of the algorithm) is the game matrix $C$.

To solve ITP more convenient by the means of many transformations [8], it is possible to transform the problem (1) - (5) to one of LPP form which is convenient for analysis and realization. For this purpose we reduce the initial number ($m \times n$) of variables by expressing ($m + n - 1$) of the basis variables through the variables left (free variables):

$$x_{11} = a_1 - \sum_{j=2}^{n} b_j + \sum_{i=2}^{m}\sum_{j=2}^{n} x_{ij}; \qquad (6)$$

$$x_{i1} = a_i - \sum_{j=2}^{n} x_{ij}, \qquad i = 2, \dots, m; \qquad (7)$$

$$x_{1j} = b_j - \sum_{i=2}^{m} x_{ij}, \qquad j = 2, \dots, n; \qquad (8)$$

Then we will get the smaller problem in which we should find not the whole matrix $X$, but its block $\tilde{X}$ (it includes all elements of the matrix $X$ except the first line and the first column), having the form of not TP but one of LPP:

$$L(X) = \sum_{i=2}^{m}\sum_{j=2}^{n} \tilde{c}_{ij} x_{ij} \qquad (9)$$

where $\tilde{c}_{ij} = c_{11} - c_{i1} - c_{1j} + c_{ij}$

$$\sum_{j=2}^{n} b_j - a_1 - \sum_{i=2}^{m}\sum_{j=2}^{n} x_{ij} \leq 0; \qquad (10)$$

$$\sum_{j=2}^{n} x_{ij} - a_i \leq 0, \qquad i = 2, \dots, m; \qquad (11)$$

$$\sum_{i=2}^{m} x_{ij} - b_j \leq 0, \qquad j = 2, \dots, n; \qquad (12)$$

$$\tilde{X}_{opt} \to \arg\max_{\tilde{X}} L(\tilde{X}) \qquad (13)$$

When we solve (9) - (13), find $(m-1) \times (n-1)$, variables and the rest of $(m + (n-1))$ variables shall be calculated according to formulas (6) - (8) that gives us the complete solving of initial TP. To solve LPP (9) – (13) we can use any of the standards methods [7] or use algorithmically simple and more predictable method based on the solving the matrix game [5] (for example, method of fictitious play) corresponding to the given LPP.

In such a way the described definition and transformation of TP to LPP gives a possibility to execute all operations of the optimal allocation of problems in any recent SRDM accurate to the current adequacy of the game matrix to the preferences of the decision-maker.

The adequacy of the game matrix $C$ to the preferences of the decision-maker is achieved by solving ITP (see step 4 of the algorithm). This can be made with the help of the inverse LPP [4] considering the described transformations. The main calculation expression corresponding to the point estimation algorithm and allowing to calculate new values of the matrix element estimations $\tilde{C} = \|\tilde{c}_{ij}\|_{mn}$ after each new observation has the following view:

$$\hat{c}_{ij}^k = \left( \sum_{j=2}^{n}\sum_{i=2}^{m} \left( \sum_{t=1}^{k} \beta^t e_{ij}^t \right)^2 \right)^{-1} \sum_{t=1}^{k} \beta^t e_{ij}^t, \qquad (14)$$

where $e_{ij}$ – components of the normal vector of unit length (NVUL) which are scaled (reduced to the unit length) vector components (matrix) of evaluations $\tilde{c}_{ij}$; $\beta$ - weight factors which reflects the informativeness of the next $k$- observation calculated as the observation vector length till its normalizing.

The specific character of TP seals on the structure of the region of feasibility (ROF) of LPP received from TP. The studies [5, 8] showed that the typical view of ROF for the problem with the size ($m = 2, n = 3$) is similar to the view showed in Figure 1 (numbers of limitations forming ROF are marked with numbers) and the character of the evaluation specification $\tilde{c}_{ij}$ (for one of the variables) from one step to another has a view showed in Figure 2.

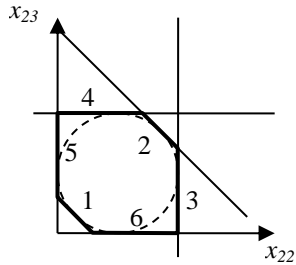 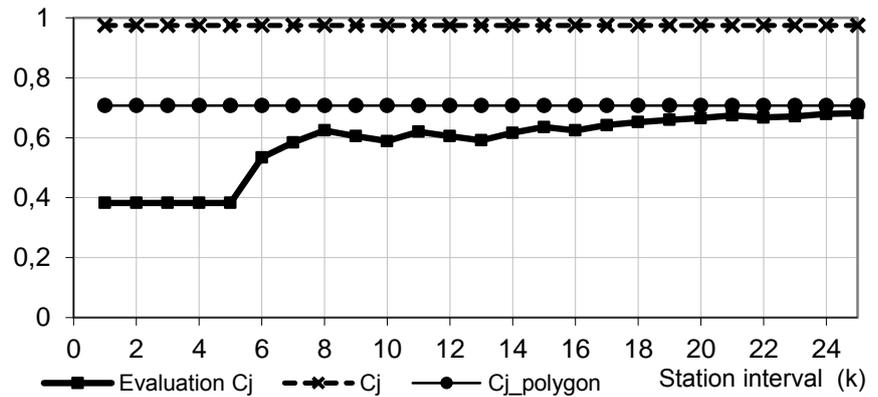

Figure 1. Typical view of TP ROF    Figure 2. Convergence of evaluations of generalized game matrix

The studies of generalized game matrix adaptation algorithm to the real preferences of the decision-maker showed that the following TP peculiarities. It should not be expected that the evaluations will be convergent to the real (simulated) values of the game matrix. Their convergence with the closest "representative" of ITP-polygon (reflecting SRDM of the highest complexity for TP of this size) is enough. At that, the quality of decisions made (plans for the current problems allocation for robots) doesn't go down that proves the efficiency of the described scheme of the robotic system adaptation to the decision-maker's preferences.

It should be noted that the rate of the evaluations convergence is not an end in itself at adaptive allocation of problems in the group of robots. The very important factor is the convergence on the problem solving. As showed in a number of works on this subject [5, 8], the convergence on the problem solving is higher than the convergence of evaluations. The convergence on the problem solving means [5] the rate of solving coincidence received upon real (simulated) game matrix and upon current evaluations. Besides the important factor which allows making the adaptation "tighter", i.e. to monitor as quick as possible the changes in the elements of the game matrix of the decision-maker, is the technology of the optimal planning of the experiments on the decision-maker [5], i.e. generation of SRDM which detect the current preferences of the decision-maker as quick as possible.

**Sources of uncertain and incomplete information during adaptive control.**

As the basic element of planning (problem allocation between robots in a group) is the preference model, represented as TP, then the information incompleteness and other forms of uncertainty may be showed up in an inaccurate parametrization of a model which shall influence on a plan quality. The causes of uncertainty factor influence may be, for example, the enemy countermeasures, communication interruptions between the robotic system and the decision-maker, communication interference, sensory field reduction of the robotic system etc.

Traditionally in the operation research problems the issue about the divergence of parameters on the optimal decision has been studied in the special section called "the analysis of the decision for the responsive". But such a way in the robotic system control is not functional. Let's discuss how the factors with the influence on the completeness of information can be considered in the robotic system control scheme.

All basic elements and parameters involved in the robotic system control are showed in Figure 3.

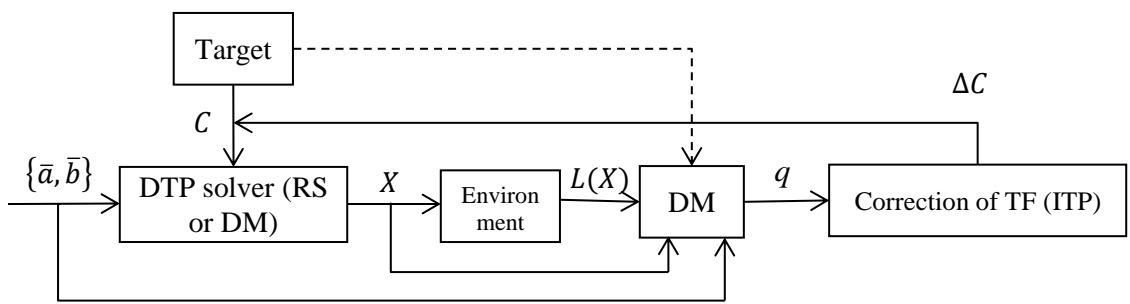

Рис. 3. Robotic system control scheme

The basic possible demonstrations of the information uncertainty and incompleteness during robotic system work in the adaptive mode are capable to influence on the plans quality.

1. When the communication session is delayed (until it stops) on the channel "robotic system – decision-maker", the correction ($\Delta C$) of the game matrix elements won't happen and this will lead to the data deterioration (actuality loss) about the target preferences of the decision-maker ($L(X)$), which means the loss of the robotic system work concerning the desired level which means that the decision quality ($q$) will be evaluated as bad by the decision-maker but it won't lead to the correction of TF.

2. The measuring characteristics loss of the robotic system sensory field and/or separate robots of a group leads to the inappropriate perception of SRDM ($\bar{a}, \bar{b}$), which in turn affects the validity of the direct LPP solving and will lead to solving (plan) which is inadequate to SRDM even if TF is actual.

3. The optimal solving $X$, which is hard realized in the environment of the robotic system work, will lead to the effect $L(X)$ which doesn't correspond to the desired level.

4. The perception of set of three elements ($\{\bar{a}, \bar{b}\}, X, L(X)$) by a decision-maker (here is more applicable to call him a decision-estimator (DE)) may contain the uncertainties (at perception) for each element and this can lead to an error in the binary classification $q$, i.e. to the errors of the first and the second kind. This in turn will lead to the situation when the bad decision is used to set TF (i.e. to form ($\Delta C$) or to refuse a good decision and not use it in settings). The studies are showing that the second case is not so essential, but the first case may lead to the relevant deformation of TF and to the bad quality of the following decisions (plans) made based on the current evaluation of the game matrix (TF).

5. Block "TF correction" realized the solving algorithm of the inverse TP and only the uncertainty of inability of $\Delta C$ to run to "Solver of the direct TP" can be connected to the block (see p.1).

6. Another important thing about the uncertainty is related to the target setting (this relation is showed with the dashed line in Figure 3). In general cases a supervisor is above of any decision-maker. The supervisor generates and brings to the decision-maker a target with which he measures the robotic system operation in the environment. If the decision-maker is not sensitive enough to the target, corrupts it accidentally or deliberately, then the supervisor will find out at some stage that the decision-maker doesn't perceive its target adequately, for example, at the stage of targets monitoring, the same way as the decision-maker does it, i.e. with the help of the same set of elements ($\{\bar{a}, \bar{b}\}, X, L(X)$). This is one of the variants of the internal control which is used in the control systems to correct the control procedures.

The described list of situations and sources of uncertainties in the scheme of the robotic system adaptive control is large and requires the future research in this area.

**Conclusion**

A wide range of models which can be used in the area of the robotic system control are the models of allocation and assignment. They can be successfully represented by the scheme of the classical transportation problem. But a priori and current criterion uncertainty related to many optimization models at the attempt of their standard use complicates their use in practice. The scheme and algorithms of the transportation problem use in the adaptive form which allow reducing the criterion uncertainty by including in the adaptive cycle a decision-maker as the target definer is described in this work.

A lot of sources of uncertainty exist in the adaptive scheme of the transportation problem solving for the targets of problems allocation in a group of robots. The record and the further analysis of the relevance of the sources of uncertainty allow building up an effective and fail-active system of the robotic system control.

**Reference**

1. Zhdanova A.A. Autonomous artificial intelligence. – M.: BINOM, 2008, 359 p.
2. Tripathi G.N., Rihani V. Motion planning of an autonomous mobile robot using artificial neural network. - www.arxiv.org.
3. Kalyaev I.A., Kapustyan S.G. Problems of robots group control. // Mechatronics, automation, control, No. 6, 2009, P. 33-40.
4. Vilisov V.Ya. About algorithms of robots adaptation to target preferences of decision-maker. // Collection of lectures of All-Russia scientific and technical conference named "Extreme robotic engineering", Saint-Petersburg: Politekhnika-Servis, 2012, P. 120-126
5. Vilisov V.Ya. Adaptive choice of control solutions. Models of operation research as the method for storage of knowledge of the decision-maker. – Saarbrucken (Germany).: LAP LAMBERT Academic Publishing, 2011, 376 p.
6. Ivchenko V.D., Korneev A.A. Analysis of procedures for work breakdown in control task of robots group // Mechatronics, automation, control, No.7, 2009, P. 36-42.
7. Taha H.A. Operation research: An Introduction: Translation from English. - M.: Publishing house Viliams, 2005, 912 p.
8. Vilisov V.Ya. Transport model approximate to decision-maker's preferences // Application informatics, No. 6(30), 2010, P.101-109.